%% file: main.tex
\documentclass[runningheads]{llncs}

\usepackage{accv}

\usepackage{accvabbrv}

\usepackage{graphicx}
\usepackage{booktabs}

\usepackage{multirow}
\usepackage{array}

\usepackage[accsupp]{axessibility}  

\usepackage{hyperref}

\usepackage{orcidlink}

\begin{document}

\title{Image Deraining with Frequency-Enhanced State Space Model} 


\author{Shugo Yamashita\thanks{Corresponding author}\orcidlink{0009-0003-8817-1057} \and
Masaaki Ikehara \orcidlink{0000-0003-3461-1507}}

\authorrunning{S.~Yamashita and M.~Ikehara.}

\institute{Keio University, Yokohama, Japan\\
\email{\{yamashita,ikehara\}@tkhm.elec.keio.ac.jp}}

\maketitle

\begin{abstract}
Removing rain degradations in images is recognized as a significant issue. In this field, deep learning-based approaches, such as Convolutional Neural Networks (CNNs) and Transformers, have succeeded. Recently, State Space Models (SSMs) have exhibited superior performance across various tasks in both natural language processing and image processing due to their ability to model long-range dependencies. This study introduces SSM to image deraining with deraining-specific enhancements and proposes a Deraining Frequency-Enhanced State Space Model (DFSSM). To effectively remove rain streaks, which produce high-intensity frequency components in specific directions, we employ frequency domain processing concurrently with SSM. Additionally, we develop a novel mixed-scale gated-convolutional block, which uses convolutions with multiple kernel sizes to capture various scale degradations effectively and integrates a gating mechanism to manage the flow of information. Finally, experiments on synthetic and real-world rainy image datasets show that our method surpasses state-of-the-art methods. Code is available at \url{https://github.com/ShugoYamashita/DFSSM}.
\end{abstract}

\section{Introduction}
\label{sec:intro}
Images captured under rainy conditions suffer from deteriorated visual quality. Rain-related degradations adversely affect the performance of downstream tasks, such as object detection, object tracking, and segmentation, as well as the overall effectiveness of vision applications that employ these algorithms. Removing rain degradations in images enhances both the performance and reliability of vision systems.

Single image deraining has been extensively studied for its usefulness. Recently, deep learning-based approaches have significantly advanced its performance.
Convolutional Neural Network (CNN)-based methods~\cite{fu2017removing,li2018recurrent,ren2019progressive,jiang2020multi,wang2020dcsfn,zamir2021multi,wang2020model,fu2021rain,yi2021structure} surpassed the previous handcrafted prior methods~\cite{kang2011automatic,luo2015removing,li2016rain,zhang2017convolutional,gu2017joint}. 
However, CNNs cannot capture global features because they collect features from local regions using fixed-size kernels.
Subsequently developed, Transformer~\cite{vaswani2017attention} employs the Self-Attention (SA) mechanism that captures global information and enables inference based on input features. 
In SA, the key-query dot-product requires computational complexity of $\mathcal{O}(H^2W^2)$ for an input image with $H \times W$ pixels. Consequently, applying standard SA to image deraining in high-resolution images is computationally expensive and impractical. To alleviate this issue, SA is calculated across channels instead of spatial dimension\cite{restormer}, within localized window regions~\cite{IDT}, or sparsely~\cite{chen2023learning,chen2023sparse}. However, these methods do not necessarily achieve the effectiveness of standard SA in capturing global features.

Recently, State Space Models (SSMs)~\cite{gu2022efficiently,mamba} have emerged as innovative architectures for deep neural networks. SSMs conduct sequence-to-sequence transformation using hidden states to interact with previously scanned elements, thereby modeling long-range dependencies with linear complexity. Mamba~\cite{mamba} enhances this architecture by treating parameters of SSMs as functions of the input, realizing content-driven inference. Mamba~\cite{mamba} architecture, initially designed for one-dimensional data in natural language processing, is applied for image processing by flattening two-dimensional images into one-dimensional sequences. Its effectiveness has been demonstrated in image recognition~\cite{zhu2024vision,vmamba,huang2024localmamba,pei2024efficientvmamba}, medical image segmentation~\cite{ma2024u,liu2024swin}, and image restoration~\cite{mambair,shi2024vmambair}. 
However, in image deraining task, MambaIR~\cite{mambair}, a generalized and simple SSM, is inferior to deraining-specific models.

\input{figure_tex/spectra}
In this paper, we present an effective deraining-specific SSM, taking into account the unique characteristics of this task. We develop a novel Deraining Frequency-Enhanced State Space Model (DFSSM) by integrating the SSM with frequency domain processing. To build an efficient deraining model capable of capturing long-range dependencies, we use the SSM as the primary architecture. In addition, we adopt frequency domain processing that are effective for deraining.
As depicted in \cref{Fig.spectrum}, Fourier transformation highlights the directional structures of rain streaks. In the spatial domain, rain streaks are dispersed across the image, but in the frequency domain, they appear as concentrated components along specific directions. Thus, frequency domain processing enables the effective removal of rain streaks as a unified spectral element.

Specifically, we propose a Frequency-Enhanced State Space Block (FSSB), which uses a Fast Fourier Transform Module (FFTM) for frequency domain processing in parallel with a Vision State Space Module (VSSM) for SSM operations. 
During training, a Frequency Reconstruction loss~\cite{tu2022maxim,mao2023intriguing} is adopted to promote learning based on frequency components. 
Furthermore, a Mixed-Scale Gated-Convolutional Block (MGCB) is used to capture local features. Convolutions with multiple kernel sizes are employed to process diverse scales of rain streaks effectively. Concurrently, adopting a gating mechanism enables the selective transmission of features.

Comprehensive experiments are conducted to evaluate the proposed DFSSM. 
Across synthetic rainy image datasets such as Rain200H~\cite{Rain200HL}, Rain200L~\cite{Rain200HL}, and DID-Data~\cite{DID_Data}, as well as real-world datasets including SPA-Data~\cite{wang2019spatial} and LHP-Rain~\cite{LHP_Rain}, our method achieves quantitatively and qualitatively superior performance compared with state-of-the-art methods. 
Ablation studies demonstrate that the proposed improvements are effective in removing rain degradations.

The contributions of this paper are summarized as follows:
\begin{itemize}
\item 
We propose a Deraining Frequency-Enhanced State Space Model (DFSSM). Our method uses a Frequency-Enhanced State Space Block (FSSB) containing a Vision State Space Module (VSSM) and a Fast Fourier Transform Module (FFTM) for effective image deraining.
\item 
We develop a Mixed-Scale Gated-Convolutional Block (MGCB), which effectively captures various scale degradations through mixed-scale convolutions and manages the flow of features using a gating mechanism.
\item 
Experimental results show that our method achieves state-of-the-art performance on synthetic and real-world rainy image datasets.
\end{itemize}

\section{Related Works}
\subsection{Image Deraining}
Extensive research has been conducted to restore rain-degraded images. Traditional methods~\cite{kang2011automatic,luo2015removing,li2016rain,zhang2017convolutional,gu2017joint} separate a rainy image into the background and rain layers with various handcrafted priors, such as discriminative sparse coding~\cite{luo2015removing} and Gaussian mixture model~\cite{li2016rain}. However, these prior-based approaches have difficulty dealing with complex rain forms. 

Currently, deep learning-based methods are the dominant approaches for image deraining.
Initially, CNN-based methods~\cite{fu2017removing,li2018recurrent,ren2019progressive,jiang2020multi,wang2020dcsfn,zamir2021multi,wang2020model,fu2021rain,yi2021structure} were extensively studied. DDN~\cite{fu2017removing} used a deep convolutional neural network, focusing on high-frequency detail during training. 
Recursive computation~\cite{li2018recurrent,ren2019progressive,jiang2020multi,wang2020dcsfn} and multi-scale representation~\cite{jiang2020multi,wang2020dcsfn,zamir2021multi} were successfully deployed to 
remove rain streaks. 
RCDNet~\cite{wang2020model} proposed interpretable network architecture with a convolutional dictionary learning model. 
DualGCN~\cite{fu2021rain} used dual graph convolutional networks to consider long-range contextual information. 
SPDNet~\cite{yi2021structure} employed a wavelet-based feature extraction backbone with a residue channel prior guided mechanism.

Recent studies~\cite{restormer,IDT,chen2023learning,chen2023sparse} have applied Transformer~\cite{vaswani2017attention} to image deraining. 
These methods are designed to reduce the computational quadratic complexity of standard Self-Attention. 
Restormer~\cite{restormer} applied Self-Attention across channels instead of the spatial dimension. IDT~\cite{IDT} used a window-based Transformer and a spatial Transformer to extract local and global features. A sparse attention mechanism, which calculates attention using only highly relevant elements rather than all elements, is introduced to rain removal along with a learnable top-k selection operator~\cite{chen2023learning} and an uncertainty-driven ranking strategy~\cite{chen2023sparse}.

\subsection{State Space Models}
State Space Models (SSMs), recently incorporated into deep learning from classical control theory, are designed for sequence-to-sequence transformation. Structured State Spaces model (S4)~\cite{gu2022efficiently} handles long sequences effectively with linear computation. Mamba~\cite{mamba} enabled content-driven inference with parameters of SSMs as functions of the input and achieved state-of-the-art performance in the natural language domain. Vim~\cite{zhu2024vision} and VMamba~\cite{vmamba}, which adapted Mamba to the field of image recognition, transform 2D images into 1D sequences by scanning with multiple routes and then using SSMs. LocalMamba~\cite{huang2024localmamba} improves the scan direction and enhances the ability to capture local 2D relationships. EfficientVMamba~\cite{pei2024efficientvmamba} proposed a lightweight model. SSMs have also been applied in medical image segmentation~\cite{ma2024u,liu2024swin} and image restoration~\cite{mambair,shi2024vmambair}. In this study, we utilize SSM with frequency enhancement to remove rain degradations.

\section{Proposed Method}
\input{figure_tex/overview}

\subsection{Overall Pipeline}
The overall architecture of the proposed Deraining Frequency-Enhanced State Space Model (DFSSM) is illustrated in \cref{Fig.overview}. Given a rainy input image $I \in \mathbb{R}^{3 \times H \times W}$, our deraining network removes degradations and outputs a clear image $\hat{I} \in \mathbb{R}^{3 \times H \times W}$. Our hierarchical encoder-decoder network employs SSM with MambaIR~\cite{mambair} as the baseline. 

We employ a U-Net~\cite{U-Net} architecture akin to Restormer~\cite{restormer}, comprising $8$ stages across four resolution levels: ${H}\times{W}$, $\frac{H}{2}\times\frac{W}{2}$, $\frac{H}{4}\times\frac{W}{4}$ and $\frac{H}{8}\times\frac{W}{8}$. Initially, the input image $I$ undergoes channel expansion from $3$ to $C$ via a $3 \times 3$ convolution. Then, the encoder and decoder are applied, processing features while decreasing and increasing resolution, respectively.
Downsampling and upsampling of image sizes are conducted through pixel-unshuffle and pixel-shuffle techniques~\cite{shi2016real}. 
Skip connections facilitate the integration of features from the encoder into the decoder. After the skip connections at $\frac{H}{2}\times\frac{W}{2}$ and $\frac{H}{4}\times\frac{W}{4}$ resolution levels, the channels are halved through a $1\times1$ convolution. 
The features from the decoder are fed into a refinement stage operating at high resolution. Finally, applying a $3 \times 3$ convolution and adding skip connection from the input image $I$, we obtain the clear image $\hat{I}$.

Each encoder, decoder, and refinement stage comprises $N_{S}$ State Space Groups (SSGs) and $N_{F}$ Frequency-Enhanced State Space Groups (FSSGs). The SSG is a series connection of a State Space Block (SSB) and a Mixed-Scale Gated-Convolutional Block (MGCB). The FSSG is a series connection of a Frequency-Enhanced State Space Block (FSSB) and an MGCB.
The following describes the components and loss functions of our model in detail.

\subsection{State Space Block (SSB)}
We employ the State Space Model (SSM) to remove rain degradations. The SSM can capture global information with linear complexity. Following MambaIR~\cite{mambair}, the State Space Block (SSB), shown in \cref{Fig.overview}(c), contains a layer normalization, a Vision State Space Module (VSSM), and a skip connection using a learnable scale factor. Given the input tensor $X$, the SSB can be defined as:
\begin{equation}
X_{out} = \text{VSSM}(\text{LN}(X)) + s \cdot X,
\end{equation}
where LN represents the layer normalization and $s$ denotes the learnable scale factor. In the VSSM~\cite{vmamba} (\cref{Fig.overview}(e)), the 2D Selective Scan Module~\cite{vmamba,mambair} scans 2D feature maps in four distinct directions, flatting them into 1D sequences. Then, each sequence is processed by SSM, and the outcomes are merged.

\subsection{Frequency-Enhanced State Space Block (FSSB)}
To enhance the capability of removing rain streaks, we utilize frequency domain processing.
Transforming an image into the frequency domain allows the separation of high-frequency signals, representing fine details and textures, and low-frequency signals, corresponding to flat regions. Separately processing these distinct features within the frequency domain enhances the model expressiveness, as successfully demonstrated in various computer vision tasks~\cite{mao2023intriguing,chi2020fast,zhang2022swinfir}, including image deraining~\cite{kang2011automatic,guo2022exploring,song2023image}.
The efficacy of frequency domain processing, particularly in deraining, is supported by the following observation.
\cref{Fig.spectrum} shows 2D Fourier amplitude spectra of a rain image, a clear image, and their difference image.
Rain streaks, which typically exhibit strong structural components in specific directions, are clearly highlighted by Fourier transformation.
While rain streaks are scattered across various positions in the spatial domain, they manifest as concentrated spectral components in particular orientations in the frequency domain. Treating rain streaks as a unified spectral component facilitates effective deraining.

Motivated by these insights, we introduce the Frequency-Enhanced State Space Block (FSSB), which expands the SSB. As shown in \cref{Fig.overview}(d), the FSSB uses a Fast Fourier Transform Module (FFTM) in parallel with the VSSM: 
\begin{equation}
X_{out} = \text{VSSM}(\text{LN}(X)) + \text{FFTM}(\text{LN}(X)) + s \cdot X.
\end{equation}

In the FFTM (\cref{Fig.overview}(f)), we initially use a $1\times1$ convolution that halves the channel and a SiLU activation function~\cite{silu} to obtain the suitable feature map for processing in the frequency domain. Subsequently, a Fast Fourier Transform (FFT) transforms the feature map from the spatial domain to the frequency domain. Given that the image features are represented as a real tensor, a 2D Real FFT is used to reduce computational complexity by exploiting symmetry. To extract features in the frequency domain, we use a $1\times1$  convolution with a SiLU. Then, a 2D Real Inverted FFT (IFFT) is used to bring it back from the frequency domain to the spatial domain. Finally, a $1\times1$ convolution expands the channels to the original number of channels. Given the input features $Z$, the FFTM process is defined as:
\begin{equation}
\hat{Z} = \text{SiLU}(1\times1\text{ Conv}(Z)),
\end{equation}
\begin{equation}
Z_f = \mathcal{F}^{-1}(\text{SiLU}(1\times1\text{ Conv}(\mathcal{F}(\hat{Z})))),
\end{equation}
\begin{equation}
Z_{out} = 1\times1\text{ Conv}(Z_f),
\end{equation}
where $\mathcal{F}$ and $\mathcal{F}^{-1}$ denote the 2D Real FFT and IFFT.

\subsection{Mixed-Scale Gated-Convolutional Block (MGCB)}
SSB and FSSB can capture the spatial long-term dependency. After these components, we use the Mixed-Scale Gated-Convolutional Block (MGCB) for its proficiency in gathering local details. MGCB contains depth-wise convolutions, aggregating information from spatially neighboring pixels. To remove rain degradations of various scales, multi-scale representation~\cite{jiang2020multi,wang2020dcsfn,zamir2021multi} and convolutions~\cite{chen2023learning} are effective. Hence, we adopt multiple kernel sizes for depthwise-convolution. Following~\cite{restormer}, this convolutional block incorporates a gating mechanism. This gating mechanism allows the passage of crucial information while blocking non-essential information, thereby managing the flow of information at each hierarchical level.

Specifically, the MGCB is depicted in \cref{Fig.overview}(g). The input tensor $X$ is applied a layer normalization and fed into three parallel branches: a gate branch, a $3\times3$ dconv branch, and a $5\times5$ dconv branch. Within these branches, the channels are initially expanded by factors of $\gamma$, $\frac{\gamma}{2}$, and $\frac{\gamma}{2}$, respectively, through $1\times1$ convolutions. $\gamma$ is a hyperparameter of the expansion rate. In the gate branch, a $3\times3$ depth-wise convolution followed by a GELU activation function~\cite{gelu} is conducted to get $X_{gate}$. In the $3\times3$ and $5\times5$ dconv branches, $3\times3$ and $5\times5$ depth-wise convolutions are performed with output $X_{3\times3}$ and $X_{5\times5}$, respectively.
In the process of integrating these branches, $X_{3\times3}$ and $X_{5\times5}$ are concatenated on a channel-wise basis, subsequently taking the element-wise product with $X_{gate}$. Then, a $1\times1$ convolution is conducted to restore the channel dimension. Furthermore, we use the channel attention~\cite{senet} to consider the global context in this block. Last, a skip connection with a learnable scale factor is added to obtain $X_{out}$. In conclusion, the MGCB can be formulated as:
\begin{equation}
X_{gate} = \text{GELU}(3\times3~\text{DW-Conv}(\text{LN}(X))),
\end{equation}
\begin{equation}
X_{3\times3} = 3\times3~\text{DW-Conv}(\text{LN}(X)),
\end{equation}
\begin{equation}
X_{5\times5} = 5\times5~\text{DW-Conv}(\text{LN}(X)),
\end{equation}
\begin{equation}
X_{out} = \text{CA}(X_{gate}{\odot}[X_{3\times3}, X_{5\times5}]) + {s}\cdot{X},
\label{eq:mgcb1}
\end{equation}
where $\text{DW-Conv}$ denotes depth-wise convolution, $\text{CA}$ stands for the channel attention, and $[\cdot]$ represents the channel-wise concatenation.

\subsection{Loss Functions}
The loss function of our method is calculated based on the model output $\hat{I}$ and the ground truth $G$. We use an L1-loss and a Frequency Reconstruction loss~\cite{tu2022maxim,mao2023intriguing}. The L1-loss is formulated as:
\begin{equation}
\mathcal{L}_{L1} = {\|\hat{I} - G\|}_1,
\label{eq:L1Loss}
\end{equation}
where ${\|\cdot\|}_1$ is the $L1$ norm.
The Frequency Reconstruction loss~\cite{tu2022maxim,mao2023intriguing} computes the loss at each frequency component: 
\begin{equation}
\mathcal{L}_{Freq} = \|\mathcal{F}(\hat{I}) - \mathcal{F}(G)\|_1
\label{eq:FrequencyLoss}
\end{equation}
where $\mathcal{F}(\cdot)$ denotes the 2D Real Fast Fourier Transform.
The total loss is:
\begin{equation}
\mathcal{L}_{total} = \mathcal{L}_{L1} + \lambda_{f} \mathcal{L}_{Freq},
\label{eq:total_loss}
\end{equation}
where $\lambda_{f}$ are coefficients to adjust the ratio of the losses.

\section{Experiments}
\subsection{Experimental settings}
\subsubsection{Datasets.}
In our experiment, existing synthetic and real-world datasets are used to validate our proposed method. Each dataset contains image pairs; one is degraded by rain, and the other is the same scene without rain.
Rain200H and Rain200L~\cite{Rain200HL} include images of synthetic heavy and light rain streaks, respectively. Both datasets have $1,800$ training image pairs and $200$ testing image pairs. 
DID-Data~\cite{DID_Data} consists of $12,000$ synthetic training image pairs and $1,200$ testing image pairs with different rain-density levels.
SPA-Data~\cite{wang2019spatial} and LHP-Rain~\cite{LHP_Rain} are real-world rainy image datasets comprising $638,492$ and $758,997$ training image pairs, respectively. Each dataset includes $1,000$ testing image pairs.

\subsubsection{Implementation Details.}
Our proposed DFSSM is implemented on PyTorch framework~\cite{pytorch}. 
Each stage has $N_{S}=1$ SSG and $N_{F}=3$ FSSGs, as empirically determined in \cref{subsec:ablation}. 
The initial channel $C$ is 32. In the MGCB, the expansion rate~$\gamma$ is $2.0$. The coefficient $\lambda_{f}$ of the loss functions is set to $0.01$.

Our DFSSM is trained for $600,000$ iterations on LHP-Rain~\cite{LHP_Rain} and for $300,000$ iterations on the other datasets. The initial learning rate $3\times10^{-4}$ is gradually decreased to $1\times10^{-6}$ with the cosine annealing strategy~\cite{loshchilov2017sgdr}. Our model is optimized by AdamW~\cite{loshchilov2018decoupled} with $\beta_1 = 0.9$, $\beta_2 = 0.999$, and weight decay $=1\times10^{-4}$. The batch size is 4.
The patch size in training is set to $128\times128$ on Rain200H and Rain200L~\cite{Rain200HL}, whereas it is set to $256\times256$ on the other datasets. We apply random cropping along with horizontal and vertical flipping for data augmentation.

\subsubsection{Evaluation metrics.}
Model performance is evaluated using peak signal-to-noise ratio (PSNR)~\cite{psnr} and structural similarity (SSIM)~\cite{ssim}. Following previous studies~\cite{restormer,IDT,chen2023learning}, PSNR and SSIM are measured based on the luminance channel Y of the YCbCr color space.

\subsection{Comparison with state-of-the-art methods}
\input{table/compare_5datasets}
\input{figure_tex/Rain200H_10img}
\input{figure_tex/LHP_Rain_10img}

\subsubsection{Compared Methods.}
We compare our DFSSM with state-of-the-art single image deraining methods, including 
prior-based methods (DSC~\cite{luo2015removing} and GMM~\cite{li2016rain}), 
CNN-based methods (DDN~\cite{fu2017removing}, RESCAN~\cite{li2018recurrent}, PReNet~\cite{ren2019progressive}, MSPFN~\cite{jiang2020multi}, RCDNet~\cite{wang2020model}, MPRNet~\cite{zamir2021multi}, DualGCN~\cite{fu2021rain}, and SPDNet~\cite{yi2021structure}), 
and Transformer-based methods (Restormer~\cite{restormer}, IDT~\cite{IDT}, and DRSformer~\cite{chen2023learning}). 

\subsubsection{Quantitative Comparison.}
\cref{Tbl.5datasets_result} reports the quantitative results in terms of PSNR and SSIM for single image deraining. 
Our DFSSM achieves the best results across all evaluation metrics for synthetic rain and real-world rain.
In particular, our DFSSM outperforms the state-of-the-art model DRSformer~\cite{chen2023learning} in PSNR by $0.82$dB on the Rain200H~\cite{Rain200HL}, $0.58$dB on the Rain200L~\cite{Rain200HL}, $0.31$dB on the DID-Data~\cite{DID_Data}, $1.01$dB on the SPA-Data~\cite{wang2019spatial}, and $0.44$dB on the LHP-Rain~\cite{LHP_Rain}.

\subsubsection{Qualitative Comparison.}
We conduct the qualitative comparison of deraining methods.
Our proposed DFSSM achieves remarkable performance in synthetic and real-world rain scenarios.
\cref{Fig.compare_Rain200H} visualizes the deraining results of the synthetic heavy rainy image. 
CNN-based methods, including DDN~\cite{fu2017removing}, PReNet~\cite{ren2019progressive}, RCDNet~\cite{wang2020model}, and SPDNet~\cite{yi2021structure} fail to restore the background in rainy conditions, resulting in unnatural textures, particularly around windmills. Compared with Transformer-based IDT~\cite{IDT} and DRSformer~\cite{chen2023learning}, our proposed DFSSM generates a clearer image, especially in the area within the red box. 
As shown in \cref{Fig.compare_LHP_Rain}, the results on real-world images indicate that our DFSSM achieves superior deraining outputs.
Notably, our DFSSM most effectively removes the rain streaks in the sky within the red box, and the rain splashes on the ground within the blue box.

\subsubsection{Model Efficiency Comparison.}
\input{table/model_efficiency}
\cref{Tbl.efficiency} depicts the model efficiency comparison with conventional high-performing methods~\cite{yi2021structure,restormer,IDT,chen2023learning}. We calculate FLOPs using a $256\times256$ resolution image. 
To compare computational efficiency, we build a lightweight DFSSM model, DFSSM-S. In the default DFSSM, the embedding dimension is set to $C=48$, with the number of blocks per stage being $[N_S, N_F]=[1, 3]$. Meanwhile, in DFSSM-S, these hyperparameters are set to $C=32$, $[N_S, N_F]=[1, 2]$.
Although DFSSM demonstrates superior performance compared with existing methods, it entails high FLOPs. Conversely, DFSSM-S is an efficient model that achieves competitive performance with relatively fewer parameters and lower FLOPs.

\subsection{Ablation study}
\label{subsec:ablation}

\input{table/ablation}
\input{table/SSG_FSSG}
\input{table/ablation_fftm}
\input{table/ablation_MGCB}

In ablation studies, we train all models using the Rain200H~\cite{Rain200HL} dataset under the same conditions.

\subsubsection{Ablation study of the overall architecture.}
To evaluate the components of the proposed DFSSM, we conduct an ablation study across five settings. This result is presented in \cref{Tbl.ablation}.
We adopt MambaIR~\cite{mambair}, a generalized and simple SSM, as the baseline. In setting (a), the baseline performs worse than deraining-specific Transformers in \cref{Tbl.5datasets_result}.
The baseline uses only the L1-Loss in its loss function, while setting (b) also uses Frequency Reconstruction Loss (FreqLoss)~\cite{tu2022maxim,mao2023intriguing}, resulting in improved accuracy. 
In setting (c), integrating FSSB into the setting (b) to incorporate frequency domain processing enhances the efficacy of rain removal. Additionally, setting (d) alters the convolutional block in the baseline to the MGCB, enabling it to effectively address multi-scale rain degradations and manage the flow of information, thereby improving performance. Setting (e), including all these improvements, surpasses the baseline with a gain of $0.99$~dB in PSNR and $0.013$ in SSIM.

\subsubsection{Varying the Number of SSG and FSSG.}
We develop the FSSB by incorporating FFTM in parallel with VSSM in the SSB. The SSG contains the SSB, while the FSSG contains the FSSB. Each encoder, decoder, and refinement stage has $N_{S}$ SSGs and $N_{F}$ FSSGs.
\cref{Tbl.SSG_FSSG} demonstrates the impact of varying the number of SSGs and FSSGs while maintaining their total count at $4$. Increasing the number of FSSGs enhances performance, but increasing from $3$ to $4$ FSSGs does not yield further improvement. Thus, in the default setting, each stage comprises $N_{S}=1$ SSG and $N_{F}=3$ FSSGs.

\subsubsection{Ablation study of FFTM.}
\cref{Tbl.ablation_fftm} presents the results of an ablation study concerning the configuration of the Fast Fourier Transform Module (FFTM). Initially, to verify the effectiveness of adopting FFT, we compare it against a module without FFT. The results indicate that removing FFT degrades performance, confirming that transforming and processing in the frequency domain via FFT is effective for image deraining. Likewise, removing the spatial domain processing by $1\times1$ convolution with SiLU activation function before the FFT and $1\times1$ convolution after the IFFT deteriorates performance. Thus, these spatial domain processes are necessary.

\subsubsection{Effectiveness of MGCB.}
We compare the Mixed-Scale Gated-Convolutional Block (MGCB) with existing convolutional blocks. 
In the baseline~\cite{mambair}, the convolution layer (ConvLayer) comprises two $3\times3$ convolutions, an activation function between them, and the channel attention~\cite{senet}.
The Gated-Dconv Feed-Forward Network (GDFN)~\cite{restormer} comprises a gating mechanism and two $3\times3$ depth-wise convolutions. The Mixed-Scale Feed-Forward Network (MSFN)~\cite{chen2023learning} incorporates two sets of $3\times3$ and $5\times5$ depth-wise convolutions, though it does not use a gating mechanism. Since GDFN and MSFN do not employ the channel attention, versions with the channel attention are also included in the comparison. As indicated in \cref{Tbl.ablation_MGCB}, the proposed MGCB outperforms all other methods, with slightly more parameters and FLOPs than GDFN~\cite{restormer} and ConvLayer in MambaIR~\cite{mambair}, and $3.1$M parameters and $35$~GFLOPs fewer than MSFN~\cite{chen2023learning}.
In our MGCB, the mixed-scale convolutions and the gating mechanism work effectively.

\subsubsection{Effectiveness of SSM.}
\input{table/SA_SSM}
\input{figure_tex/flops_input_size}

We evaluate the effectiveness of the SSM adopted in our DFSSM. We compare three methods with global receptive fields: State Space Model (SSM)~\cite{mamba}, standard Self-Attention (SA)~\cite{vaswani2017attention}, and Multi-Dconv Head Transposed Attention (MDTA)~\cite{restormer}, which calculates SA across channels rather than spatial dimension.
Models, where the SSM in DFSSM is replaced with SA or MDTA, are trained under the same experimental conditions. As shown in \cref{Fig.flops}, when the input image size increases, computational cost (FLOPs) of SA increases quadratically, while that of MDTA and SSM increases linearly.
\cref{Tbl.SA_SSM} shows the performance for both the $128\times128$ cropped images and the original size images. 
We use an Nvidia RTX A6000 (48GB GPU memory), but SA cannot process the original image size due to its high computational cost, resulting in out-of-memory. 
For $128\times128$ cropped images, SSM exhibits similar performance with only $6.1$\% lower computational costs compared with SA. Whether for cropped or original images, SSM achieves higher performance with slightly lower computational cost than MDTA.

\section{Conclusion}
In this paper, we present a novel Deraining Frequency-Enhanced State Space Model (DFSSM). To effectively remove rain degradations, we design a Frequency-Enhanced State Space Block (FSSB), which employs a State Space Block (SSB) to model long-range dependencies and a Fast Fourier Transform Module (FFTM) for frequency domain processing in parallel. Consequently, we propose a Mixed-Scale Gated-Convolutional Block (MGCB) to capture various scale rain degradations and control the flow of features. In synthetic and real-world rain scenarios, our proposed DFSSM performs favorably against state-of-the-art algorithms quantitatively and qualitatively.


%
%
\bibliographystyle{splncs04}
\bibliography{main}
\end{document}

%% file: figure_tex/spectra.tex
\begin{figure}[t]
\centering
\begin{minipage}[t]{.26\hsize}
\centering
\includegraphics[width=0.8\hsize]{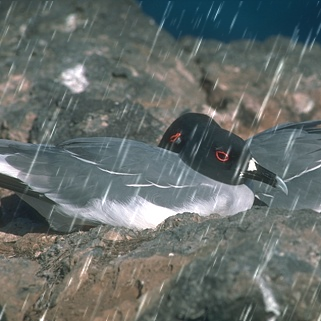}
{$\text{Rainy}$}
\end{minipage}
\begin{minipage}[t]{.26\hsize}
\centering
\includegraphics[width=0.8\hsize]{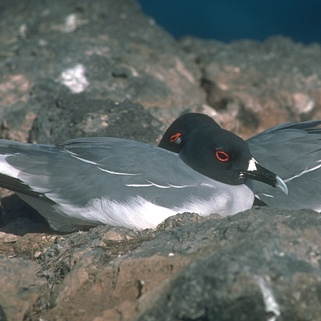}
{$\text{Clear}$}
\end{minipage}
\begin{minipage}[t]{.26\hsize}
\centering
\includegraphics[width=0.8\hsize]{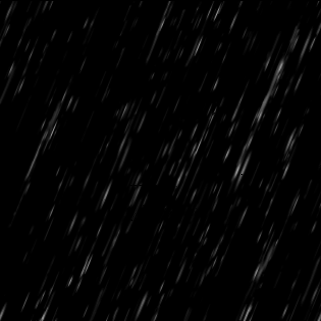}
{$\text{Rainy} - \text{Clear}$}
\end{minipage}
\begin{minipage}[t]{.26\hsize}
\centering
\includegraphics[width=0.8\hsize]{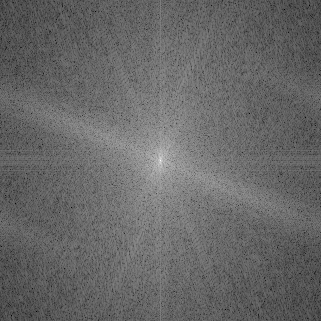}
{$|\mathcal{F}(\text{Rainy})|$}
\end{minipage}
\begin{minipage}[t]{.26\hsize}
\centering
\includegraphics[width=0.8\hsize]{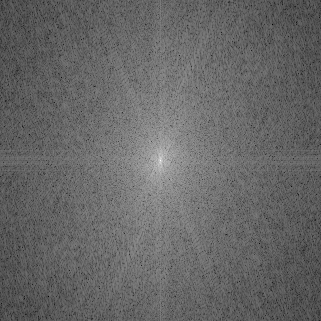}
{$|\mathcal{F}(\text{Clear})|$}
\end{minipage}
\begin{minipage}[t]{.26\hsize}
\centering
\includegraphics[width=0.8\hsize]{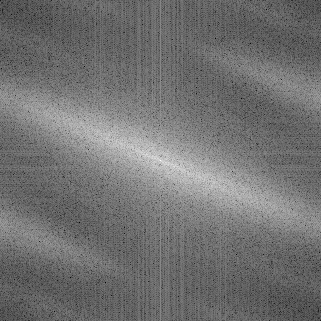}
{$|\mathcal{F}(\text{Rainy} - \text{Clear})|$}
\end{minipage}
\caption{
Top row: a rainy image, a clear image, and their differential image. Bottom row: Corresponding 2D Fourier amplitude spectra. The amplitude of the Fourier transform is denoted as $|\mathcal{F}(\cdot)|$.
}
\label{Fig.spectrum}
\end{figure}

%% file: figure_tex/overview.tex
\begin{figure}[!t]
\centering
\includegraphics[width=1\hsize]{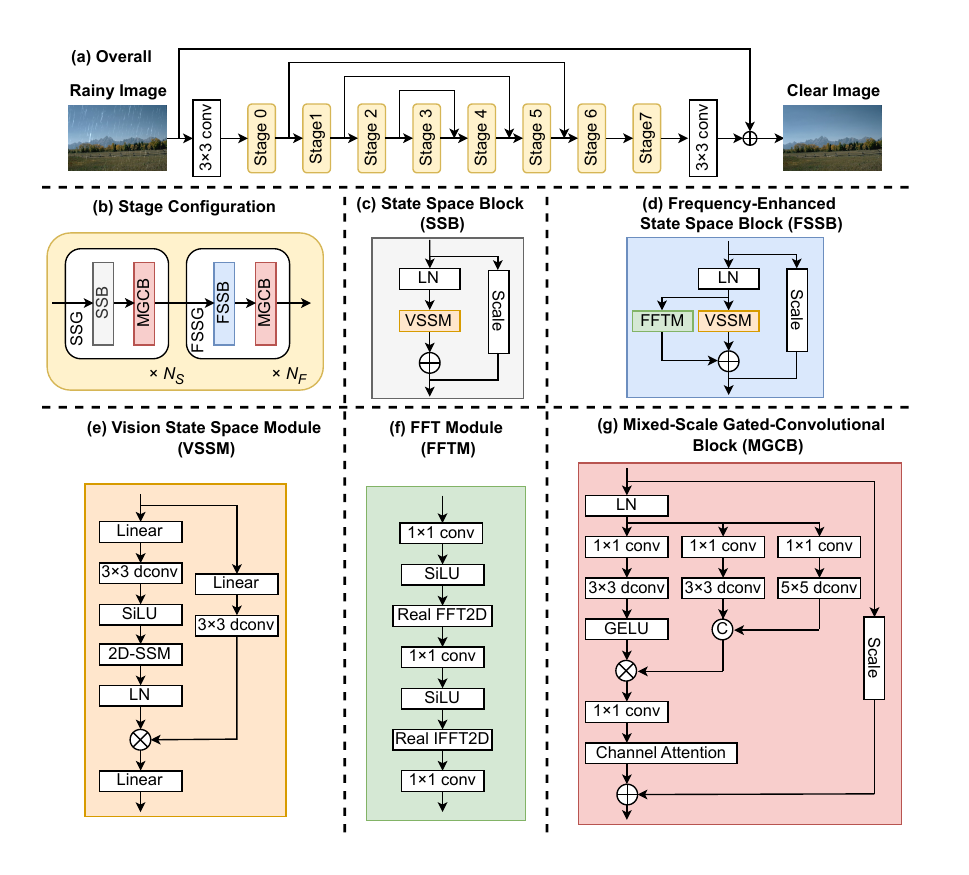}
\caption{The architecture of our Deraining Frequency-Enhanced State Space Model (DFSSM). The overall U-Net architecture (a) has $8$ stages. Each stage (b) consists of $N_{S}$ State Space Groups (SSGs) and $N_{F}$ Frequency-Enhanced State Space Groups (FSSGs). (c), (d), (e), (f), and (g) illustrate the details of the components. The SSG includes a State Space Block (SSB) and a Mixed-Scale Gated-Convolutional Block (MGCB), while the FSSG includes a Frequency-Enhanced State Space Block (FSSB) and an MGCB. Both SSB and FSSB employ a Vision State Space Module (VSSM), and the FSSB also uses a Fast Fourier Transform Module (FFTM).
}
\label{Fig.overview}
\end{figure}

%% file: table/compare_5datasets.tex
\begin{table}[t]
\centering
\caption{Quantitative results on the Rain200H, Rain200L~\cite{Rain200HL}, DID-Data~\cite{DID_Data}, SPA-Data~\cite{wang2019spatial}, and LHP-Rain~\cite{LHP_Rain}.}
\label{Tbl.5datasets_result}
\resizebox{0.99\textwidth}{!}{
\begin{tabular}{cc|cccccccccc}
\hline
\multicolumn{1}{c|}{\multirow{2}{*}{Architecture}} & \multicolumn{1}{c|}{\multirow{2}{*}{Method}}& \multicolumn{2}{c}{Rain200H~\cite{Rain200HL}}  & \multicolumn{2}{c}{Rain200L~\cite{Rain200HL}} & 
\multicolumn{2}{c}{DID-Data~\cite{DID_Data}} & \multicolumn{2}{c}{SPA-Data~\cite{wang2019spatial}} & \multicolumn{2}{c}{LHP-Rain~\cite{LHP_Rain}}\\
\multicolumn{1}{c|}{}&\multicolumn{1}{c|}{}&PSNR$\uparrow$&SSIM$\uparrow$&PSNR$\uparrow$&SSIM$\uparrow$&PSNR$\uparrow$&SSIM$\uparrow$&PSNR$\uparrow$&SSIM$\uparrow$&PSNR$\uparrow$&SSIM$\uparrow$ \\
\hline
\multicolumn{1}{c|}{\multirow{2}{*}{Prior}} & DSC~\cite{luo2015removing} & 14.73 & 0.3815 & 27.16 & 0.8663 & 24.24 & 0.8279 & 34.95 & 0.9416 & - & - \\
\multicolumn{1}{c|}{}& GMM~\cite{li2016rain}  & 14.50 & 0.4164 & 28.66 & 0.8652 & 25.81 & 0.8344 & 34.30 & 0.9428 & - & - \\
\hline 
\multicolumn{1}{c|}{\multirow{8}{*}{CNN}}  &DDN~\cite{fu2017removing} & 26.05 & 0.8056 & 34.68 & 0.9671 & 30.97 & 0.9116 & 36.16 & 0.9457 & 32.61 & 0.9037 \\
\multicolumn{1}{c|}{}&RESCAN~\cite{li2018recurrent}  & 26.75 & 0.8353 & 36.09 & 0.9697 & 33.38 & 0.9417 & 38.11 & 0.9707 &32.72 & 0.9058\\
\multicolumn{1}{c|}{}&PReNet~\cite{ren2019progressive}  & 29.37 & 0.9029 & 37.95 & 0.9816 & 33.17 & 0.9481 & 40.16 & 0.9816 & 33.17 & 0.9177 \\
\multicolumn{1}{c|}{}&MSPFN~\cite{jiang2020multi}  & 29.36 & 0.9034 & 38.58 & 0.9827 & 33.72 & 0.9550 & 43.43 & 0.9843 & 32.20 & 0.8955 \\
\multicolumn{1}{c|}{}&RCDNet~\cite{wang2020model}  & 30.24 &0.9048 & 39.17 & 0.9885 & 34.08 & 0.9532 & 43.36 & 0.9831 & 32.98 & 0.9119 \\
\multicolumn{1}{c|}{}&MPRNet~\cite{zamir2021multi} & 30.67 & 0.9110 & 39.47  & 0.9825 & 33.99 & 0.9590 &43.64 & 0.9844 & 32.43 & 0.8925\\
\multicolumn{1}{c|}{}&DualGCN~\cite{fu2021rain} & 31.15 & 0.9125 & 40.73  & 0.9886 & 34.41 & 0.9624 & 44.18 & 0.9902 & 33.38 & 0.9200 \\
\multicolumn{1}{c|}{}&SPDNet~\cite{yi2021structure} & 31.28 & 0.9207 & 40.50 & 0.9875 & 34.57 & 0.9560 & 43.20 & 0.9871 & 32.98 & 0.9122 \\         
\hline
\multicolumn{1}{c|}{\multirow{3}{*}{Transformer}}  &Restormer~\cite{restormer} & 32.00 & 0.9329 & 40.99 & 0.9890 & 35.29 & 0.9641 & 47.98& 0.9921 & \underline{33.77} & \underline{0.9290} \\
\multicolumn{1}{c|}{}  &IDT~\cite{IDT} &{32.10} &\underline{0.9344} &{40.74}  &0.9884 & 34.95 & 0.9627 & 47.34 & \underline{0.9930} & 33.43 & 0.9246 \\
\multicolumn{1}{c|}{}  &DRSformer~\cite{chen2023learning} & \underline{32.17}   & 0.9326 & \underline{41.23}  & \underline{0.9894} & \underline{35.35} & \underline{0.9646} &\underline{48.54} & 0.9924 & 33.55 & 0.9227 \\
\hline
\multicolumn{1}{c|}{\multirow{1}{*}{SSM}}  
&\textbf{DFSSM}  & \textbf{32.99} & \textbf{0.9403} & \textbf{41.81} & \textbf{0.9905} & \textbf{35.66} & \textbf{0.9671} & \textbf{49.55} & \textbf{0.9939} & \textbf{33.99} & \textbf{0.9322}\\
\hline
\end{tabular}}
\end{table}

%% file: figure_tex/Rain200H_10img.tex
\begin{figure}[t]
   \centering
   \begin{minipage}[t]{.19\hsize}
      \centering
      \includegraphics[width=1\hsize]{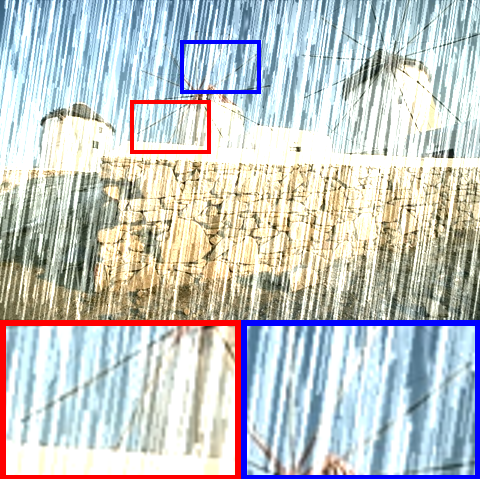}
      {\scriptsize Input}
   \end{minipage}
   \begin{minipage}[t]{.19\hsize}
      \centering
      \includegraphics[width=1\hsize]{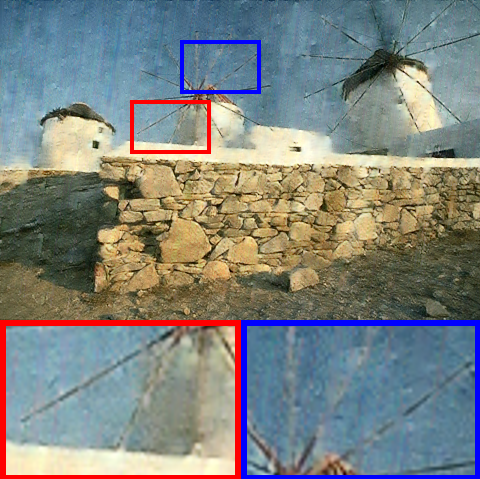}
      {\scriptsize DDN~\cite{fu2017removing}}
   \end{minipage}
   \begin{minipage}[t]{.19\hsize}
      \centering
      \includegraphics[width=1\hsize]{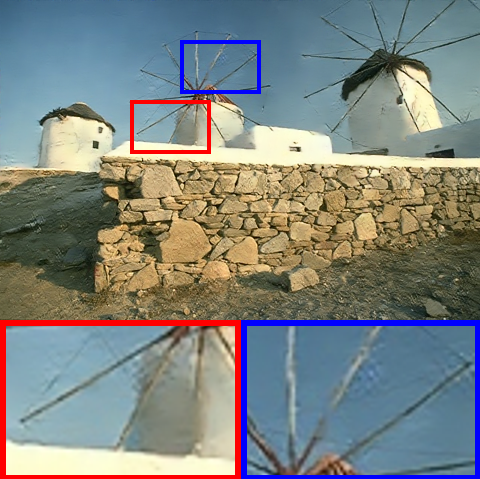}
      {\scriptsize PReNet~\cite{ren2019progressive}}
   \end{minipage}
   \begin{minipage}[t]{.19\hsize}
      \centering
      \includegraphics[width=1\hsize]{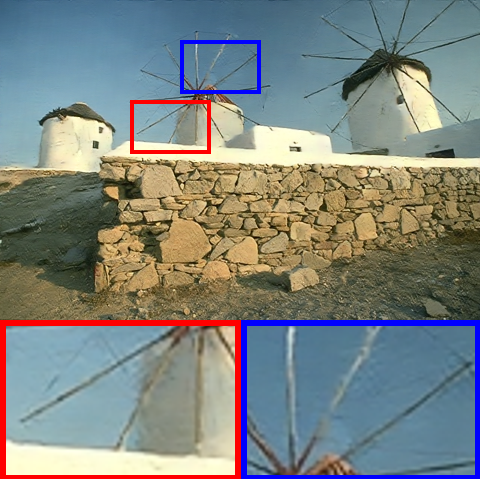}
      {\scriptsize RCDNet~\cite{wang2020model}}
   \end{minipage}
   \begin{minipage}[t]{.19\hsize}
      \centering
      \includegraphics[width=1\hsize]{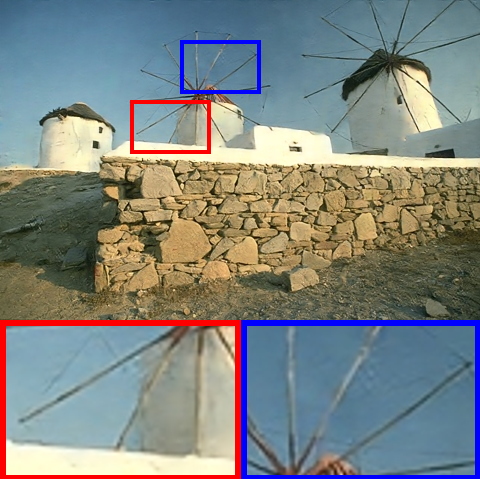}
      {\scriptsize SPDNet~\cite{yi2021structure}}
   \end{minipage}
   
   \begin{minipage}[t]{.19\hsize}
      \centering
      \includegraphics[width=1\hsize]{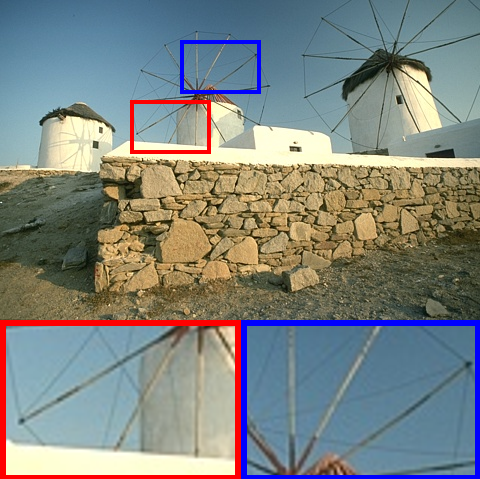}
      {\scriptsize GT}
   \end{minipage}
   \begin{minipage}[t]{.19\hsize}
      \centering
      \includegraphics[width=1\hsize]{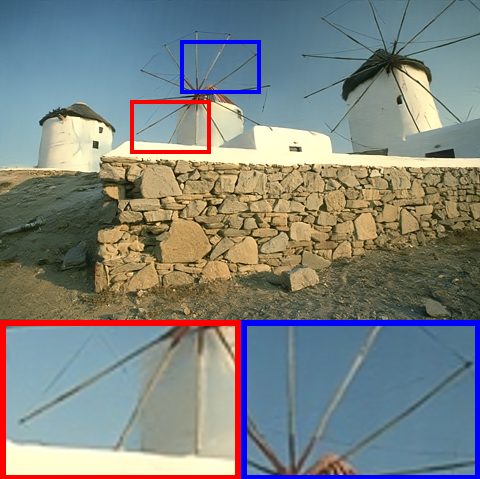}
      {\scriptsize Restormer~\cite{restormer}}
   \end{minipage}
   \begin{minipage}[t]{.19\hsize}
      \centering
      \includegraphics[width=1\hsize]{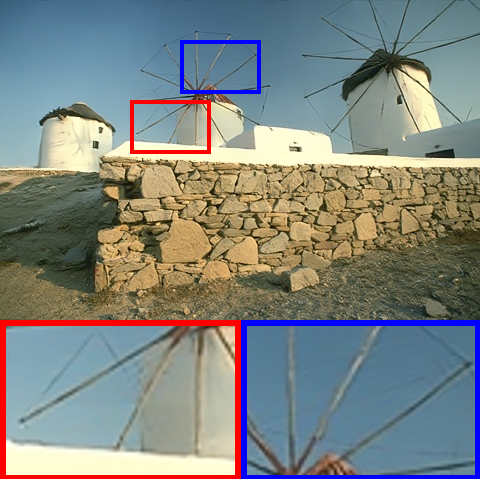}
      {\scriptsize IDT~\cite{IDT}}
   \end{minipage}
   \begin{minipage}[t]{.19\hsize}
      \centering
      \includegraphics[width=1\hsize]{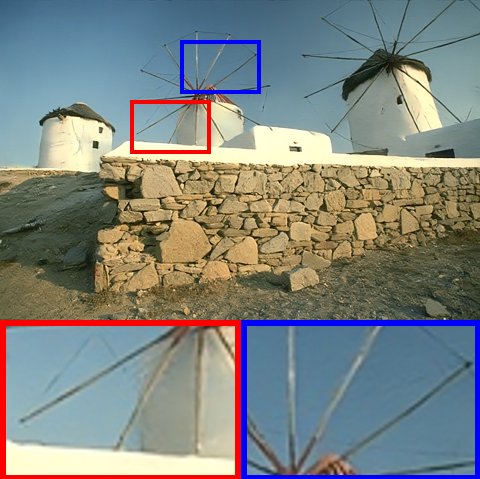}
      {\scriptsize DRSformer~\cite{chen2023learning}}
   \end{minipage}
   \begin{minipage}[t]{.19\hsize}
      \centering
      \includegraphics[width=1\hsize]{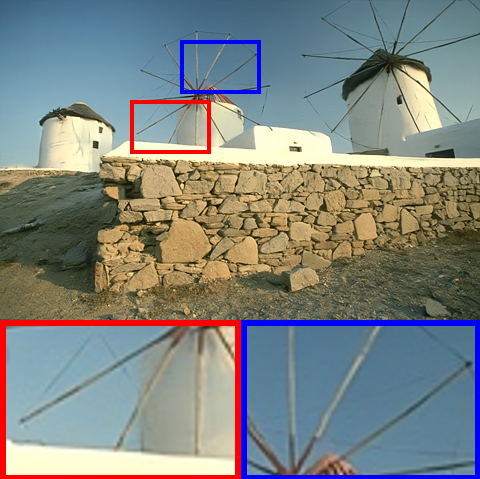}
      {\scriptsize DFSSM (Ours)}
   \end{minipage}
   \caption{
   Visual comparison on the synthetic rainy image of Rain200H~\cite{Rain200HL}. Red and blue boxes correspond to the zoomed-in patches.
}
   \label{Fig.compare_Rain200H}
\end{figure}

%% file: figure_tex/LHP_Rain_10img.tex
\begin{figure}[t]
   \centering
   \begin{minipage}[t]{.19\hsize}
      \centering
      \includegraphics[width=1\hsize]{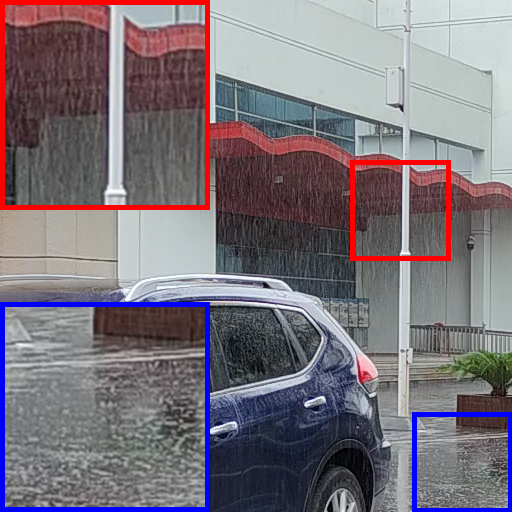}
      {\scriptsize Input}
   \end{minipage}
   \begin{minipage}[t]{.19\hsize}
      \centering
      \includegraphics[width=1\hsize]{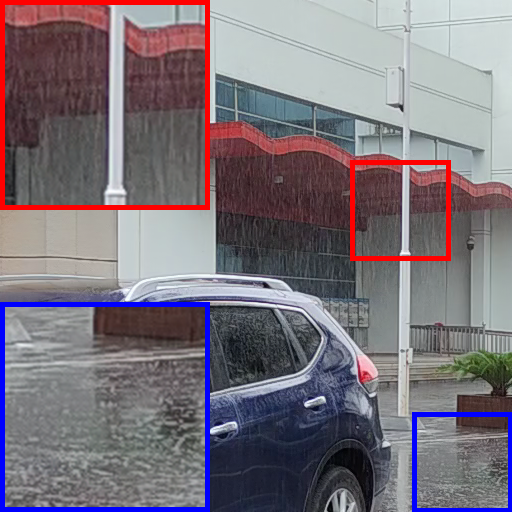}
      {\scriptsize DDN~\cite{fu2017removing}}
   \end{minipage}
   \begin{minipage}[t]{.19\hsize}
      \centering
      \includegraphics[width=1\hsize]{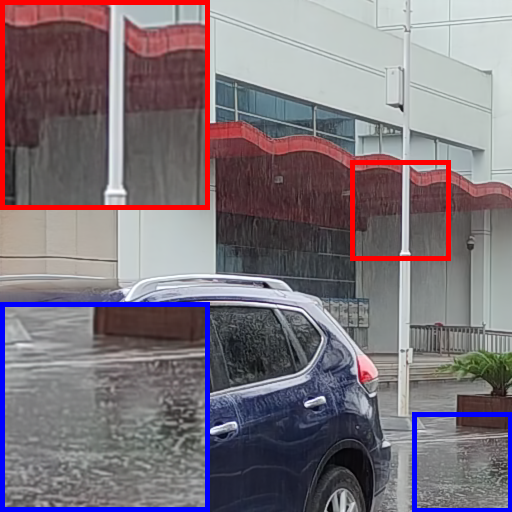}
      {\scriptsize PReNet~\cite{ren2019progressive}}
   \end{minipage}
   \begin{minipage}[t]{.19\hsize}
      \centering
      \includegraphics[width=1\hsize]{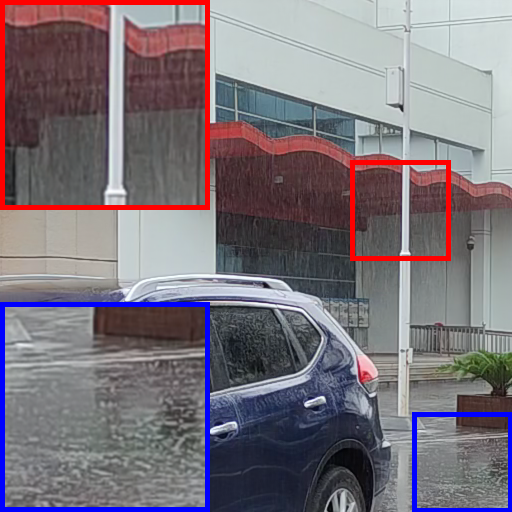}
      {\scriptsize RCDNet~\cite{wang2020model}}
   \end{minipage}
   \begin{minipage}[t]{.19\hsize}
      \centering
      \includegraphics[width=1\hsize]{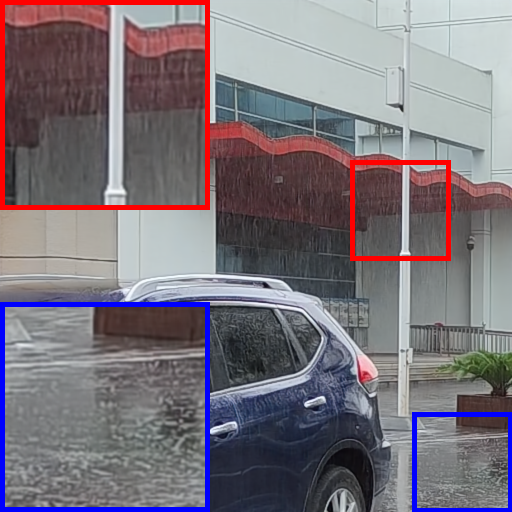}
      {\scriptsize SPDNet~\cite{yi2021structure}}
   \end{minipage}

   \begin{minipage}[t]{.19\hsize}
      \centering
      \includegraphics[width=1\hsize]{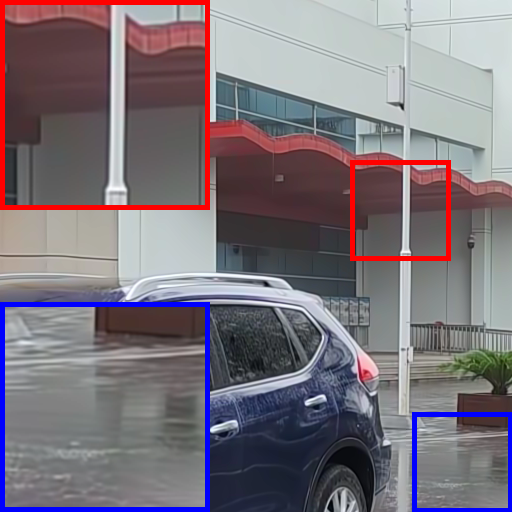}
      {\scriptsize GT}
   \end{minipage}
   \begin{minipage}[t]{.19\hsize}
      \centering
      \includegraphics[width=1\hsize]{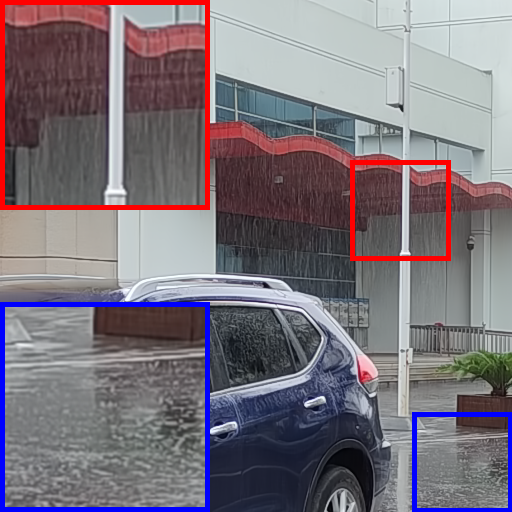}
      {\scriptsize Restormer~\cite{restormer}}
   \end{minipage}
   \begin{minipage}[t]{.19\hsize}
      \centering
      \includegraphics[width=1\hsize]{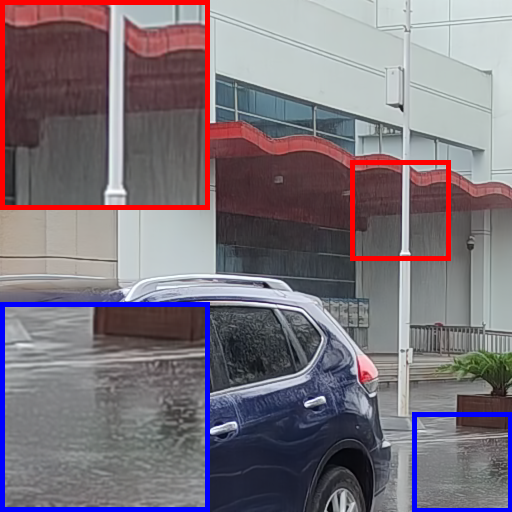}
      {\scriptsize IDT~\cite{IDT}}
   \end{minipage}
   \begin{minipage}[t]{.19\hsize}
      \centering
      \includegraphics[width=1\hsize]{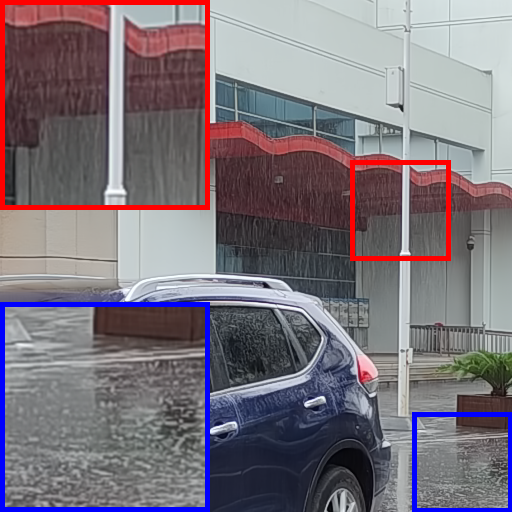}
      {\scriptsize DRSformer~\cite{chen2023learning}}
   \end{minipage}
   \begin{minipage}[t]{.19\hsize}
      \centering
      \includegraphics[width=1\hsize]{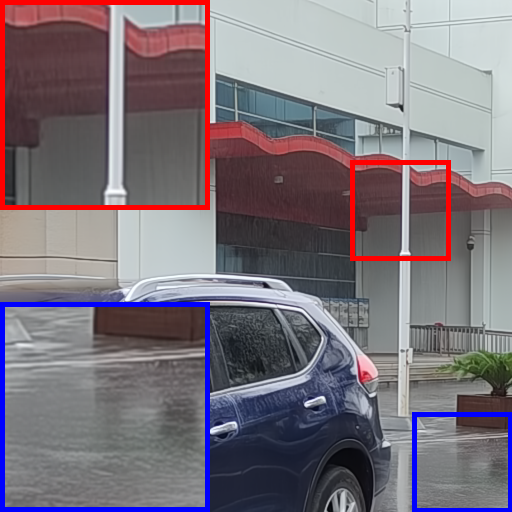}
      {\scriptsize DFSSM (Ours)}
   \end{minipage}
   \caption{
   Visual comparison on the real-world rainy image of LHP-Rain~\cite{LHP_Rain}. Red and blue boxes correspond to the zoomed-in patches.
}
   \label{Fig.compare_LHP_Rain}
\end{figure}

%% file: table/model_efficiency.tex
\begin{table}[t]
\centering
\caption{Model efficiency comparison on the Rain200H~\cite{Rain200HL}. FLOPs are calculated using a $256\times256$ resolution image.
}
\label{Tbl.efficiency}
\resizebox{0.99\textwidth}{!}{
\begin{tabular}{c|>{\centering\arraybackslash}m{2cm} >{\centering\arraybackslash}m{2cm} >{\centering\arraybackslash}m{2cm} >{\centering\arraybackslash}m{2cm} >{\centering\arraybackslash}m{2cm} >{\centering\arraybackslash}m{2cm}}
\hline
Method~&~SPDNet~\cite{yi2021structure}~&~Restormer~\cite{restormer}~&~IDT~\cite{IDT}~&~DRSformer~\cite{chen2023learning}~&~\textbf{DFSSM-S}~&~\textbf{DFSSM} \\
\hline
PSNR$\uparrow/$SSIM$\uparrow$ &$31.28/0.9207$&$32.00/0.9329$&$32.10/0.9344$&$32.17/0.9326$&$32.11/0.9313$&$32.99/0.9403$\\
Parameters (M)$\downarrow$&3.3&26.1&16.4&33.7&7.0&19.0\\
FLOPs (G)$\downarrow$&89.3&156.3&73.5&244.6&87.9&212.4\\
\hline
\end{tabular}}
\end{table}

%% file: table/ablation.tex
\begin{table}[t]
\centering
\caption{Ablation study of the overall architecture.}
\label{Tbl.ablation}
\begin{tabular}{ccccccccc}
\hline
~Setting~&~FreqLoss~&~FSSB~&~MGCB~&~PSNR$\uparrow$~&~SSIM$\uparrow$~&~Param. (M)$\downarrow$~&~FLOPs (G)$\downarrow$\\
\hline
(a) & & &  & 32.00  & 0.9273 & 17.1 & 198.3 \\
(b) & $\checkmark$ & & &  32.36 & 0.9323 & 17.1 & 198.3 \\
(c) & $\checkmark$ & $\checkmark$ & & 32.87 & 0.9385 & 18.7 & 207.9 \\
(d) & $\checkmark$ & & $\checkmark$ & 32.82 & 0.9384 & 17.4 & 202.8 \\
(e) & $\checkmark$ & $\checkmark$ & $\checkmark$ & \textbf{32.99} & \textbf{0.9403} & 19.0 & 212.4 \\
\hline
\end{tabular}
\end{table}

%% file: table/SSG_FSSG.tex
\begin{table}[t]
\centering
\caption{Varying the Number of SSGs and FSSGs based on the setting (c) (\cref{Tbl.ablation}). In the default setting, each stage comprises $1$ SSG and $3$ FSSGs.}
\label{Tbl.SSG_FSSG}
\begin{tabular}{cccccc}
\hline
~SSG~&~FSSG~&~PSNR$\uparrow$~&~SSIM$\uparrow$~&~Param.(M)$\downarrow$~&~FLOPs (G)$\downarrow$\\
\hline
4 & 0 & 32.36 & 0.9323 & 17.1 & 198.3\\
3 & 1 & 32.55 & 0.9359 & 17.7 & 201.5 \\
2 & 2 & 32.78 & 0.9379 & 18.2 & 204.7 \\
1 & 3 & \textbf{32.87} & 0.9385 & 18.7 & 207.9 \\
0 & 4 & 32.78 & \textbf{0.9386} & 19.2 & 211.1 \\
\hline
\end{tabular}
\end{table}

%% file: table/ablation_fftm.tex
\begin{table}[t]
\centering
\caption{Ablation study of our Fast Fourier Transform Module (FFTM) based on the setting (c) (\cref{Tbl.ablation}).}\label{Tbl.ablation_fftm}
\begin{tabular}{lcc}
\hline
Method~&~PSNR $\uparrow$~&~SSIM$\uparrow$~\\
\hline
\textbf{FFTM}& \textbf{32.87} & \textbf{0.9385} \\
FFTM w/o FFT & 32.58 & 0.9354 \\
FFTM w/o $1\times1$ conv in spatial domain & 32.33 & 0.9322 \\
\hline
\end{tabular}
\end{table}

%% file: table/ablation_MGCB.tex
\begin{table}[t]
\centering
\caption{Comparison of our Mixed-Scale Gated-Convolutional Block (MGCB) with previous convolutional blocks based on the setting (d) (\cref{Tbl.ablation}). CA denotes the channel attention~\cite{senet}.}
\label{Tbl.ablation_MGCB}
\begin{tabular}{lccccc}
\hline
Method~&CA~&~PSNR$\uparrow$~&~SSIM$\uparrow$~&~Param. (M)$\downarrow$~&~FLOPs (G)$\downarrow$\\
\hline
GDFN~\cite{restormer} & & 32.56 & 0.9345 & 17.2 & 201.3 \\
MSFN~\cite{chen2023learning} &  & 32.67 & 0.9363 & 20.5 & 237.7 \\
GDFN~\cite{restormer} + CA & $\checkmark$ & 32.71 & 0.9376 & 17.3 & 201.5 \\
MSFN~\cite{chen2023learning} + CA & $\checkmark$ & 32.61 & 0.9361 & 20.5 & 237.9 \\
ConvLayer in MambaIR~\cite{mambair} & $\checkmark$ & 32.36  & 0.9323 & 17.1 & 198.3 \\
\textbf{MGCB} & $\checkmark$ & \textbf{32.82} & \textbf{0.9384} & 17.4 & 202.8 \\
\hline
\end{tabular}
\end{table}

%% file: table/SA_SSM.tex
\begin{table}[t]
\centering
\caption{Comparison of State Space Model (SSM)~\cite{mamba} with Self-Attention (SA)~\cite{vaswani2017attention} and Multi-Dconv Head Transposed Attention (MDTA)~\cite{restormer}.}
\label{Tbl.SA_SSM}
\resizebox{0.99\textwidth}{!}{
\begin{tabular}{c|c|ccc|ccc}
\hline
\multicolumn{1}{c|}{\multirow{2}{*}{~Method~}} & \multicolumn{1}{c|}{\multirow{2}{*}{~Param. (M)~}}& \multicolumn{3}{c|}{~Cropped Images ($128\times128$)~}  & 
\multicolumn{3}{c}{~Original Images ($321\times421$)}~\\
\multicolumn{1}{c|}{}&\multicolumn{1}{c|}{}&PSNR$\uparrow$&SSIM$\uparrow$&FLOPs (G)$\downarrow$
&PSNR$\uparrow$&SSIM$\uparrow$&FLOPs (G)$\downarrow$ \\
\hline
SA&26.1&\textbf{31.58}&0.9288&865.2&\multicolumn{3}{c}{out of memory} \\
MDTA&26.3&31.01&0.9240&56.5&32.57&0.9356&551.4 \\
SSM&\textbf{19.0}&31.43&\textbf{0.9292}&\textbf{53.1}&\textbf{32.99}&\textbf{0.9403}&\textbf{518.8} \\
\hline
\end{tabular}}
\end{table}

%% file: figure_tex/flops_input_size.tex
\begin{figure}[!t]
\centering
\includegraphics[width=1\hsize]{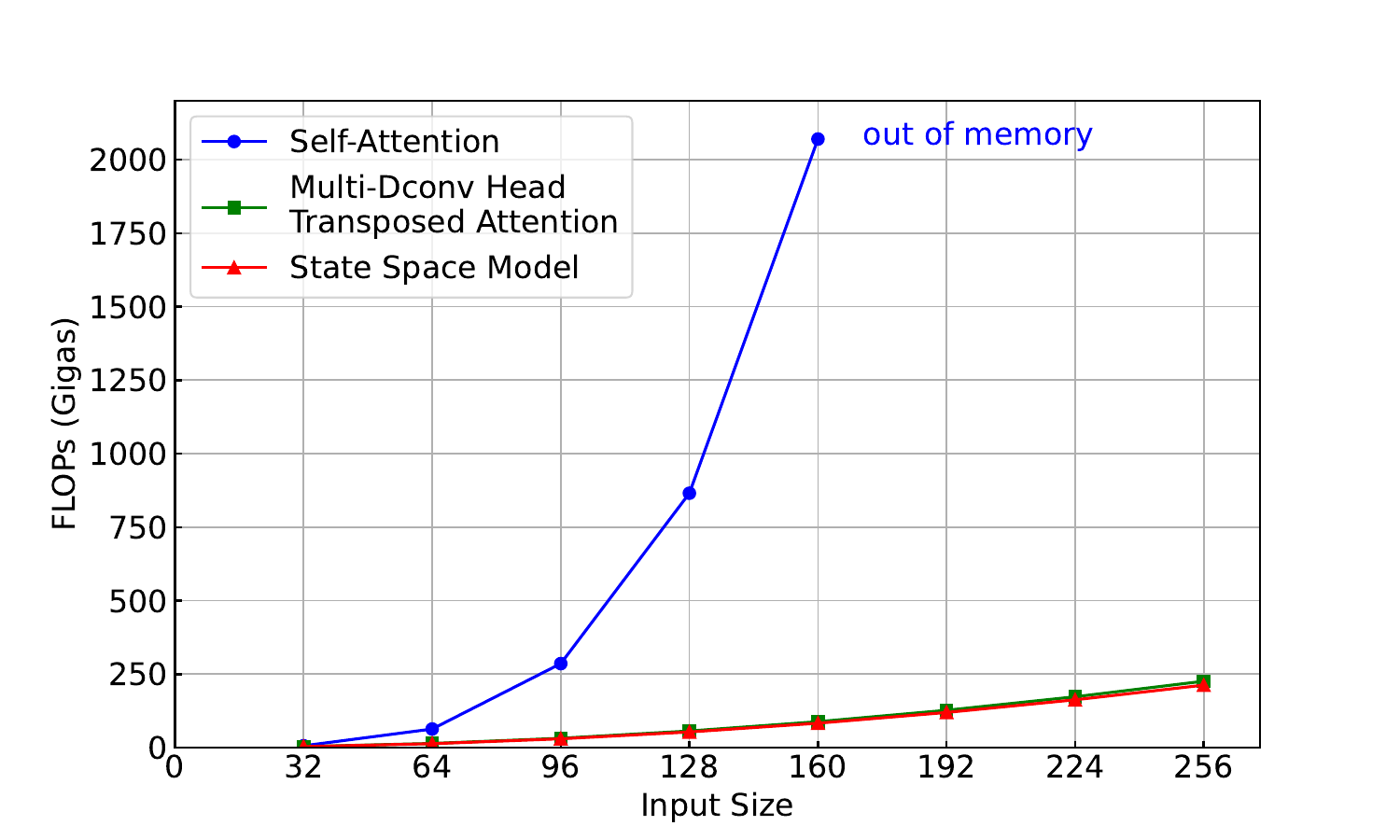}
\caption{
Comparison of computational complexity among State Space Model (SSM)~\cite{mamba}, standard Self-Attention (SA)~\cite{vaswani2017attention}, and multi-Dconv Head Transposed Attention (MDTA)~\cite{restormer}. FLOPs are measured for input image sizes ranging from $32\times32$ to $256\times256$. SA can not be measured beyond $192\times192$ due to being out of memory.
}
\label{Fig.flops}
\end{figure}